\documentclass[10pt,journal,compsoc]{IEEEtran-customize}

\ifCLASSOPTIONcompsoc
  \usepackage[nocompress]{cite}
\else
  \usepackage{cite}
\fi
\usepackage{graphicx}
\usepackage{color}
\usepackage{xcolor}
\usepackage{booktabs}
\usepackage{multirow}
\usepackage[hyphens]{url}
\usepackage{hyperref}
\usepackage{caption}
\usepackage{amsmath}
\usepackage{subcaption}
\usepackage{amsfonts, amssymb}
\usepackage{colortbl}

\definecolor{purple}{RGB}{128, 0, 128}
\definecolor{LightRed}{rgb}{1,0.92,0.92}
\definecolor{LightOrange}{rgb}{1,0.95,0.88}
\definecolor{LightYellow}{rgb}{1.0,1.0,0.84}
\definecolor{LightGreen}{rgb}{0.9,1.0,0.88}
\definecolor{LightCyan}{rgb}{0.9,1,1}
\definecolor{LightBlue}{rgb}{0.9,0.94,1}
\definecolor{LightIndigo}{rgb}{0.92,0.9,1}
\definecolor{LightMagenta}{rgb}{0.96,0.86,1}
\definecolor{DirtyWhite}{rgb}{0.96,0.96,0.96}

\DeclareSymbolFont{extraup}{U}{zavm}{m}{n}
\DeclareMathSymbol{\varheart}{\mathalpha}{extraup}{86}
\DeclareMathSymbol{\vardiamond}{\mathalpha}{extraup}{87}
\DeclareMathSymbol{\varclubsuit}{\mathalpha}{extraup}{88}

\begin{document}

\title{Recurrent Visual Feature Extraction and Stereo Attentions for CT Report Generation}

\author{Yuanhe Tian, \hspace{0.1cm}
Lei Mao, \hspace{0.1cm}
Yan Song
\thanks{
The corresponding author is Yan Song.}
\thanks{Yuanhe Tian is with University of Washington, USA (E-mail: \texttt{yhtian@uw.edu}).}
\thanks{Lei Mao is with Origin Omics, China (E-mail: \texttt{maolei@originomics-ai.com}).}
\thanks{Yan Song is with University of Science and Technology of China, China (E-mail: \texttt{clksong@gmail.com}).}
\thanks{The code is available at \url{https://github.com/synlp/R-VFE-CTRG}.}
}

\IEEEtitleabstractindextext{%

\begin{abstract}
Generating reports for computed tomography (CT) images is a challenging task, while similar to existing studies for medical image report generation,
yet has its unique characteristics, such as spatial encoding of multiple images, alignment between image volume and texts, etc.
Existing solutions typically use general 2D or 3D image processing techniques to extract features from a CT volume, where they 
firstly compress the volume and then divide the compressed CT slices into patches for visual encoding.
These approaches do not explicitly account for the transformations among CT slices, nor do they effectively integrate multi-level image features, particularly those containing specific organ lesions, to instruct CT report generation (CTRG).
In considering the strong correlation among consecutive slices in CT scans,
in this paper,
we propose a large language model (LLM) based CTRG method with recurrent visual feature extraction and stereo attentions for hierarchical feature modeling.
Specifically, we use a vision Transformer to recurrently process each slice in a CT volume,
and employ a set of attentions over the encoded slices from different perspectives to selectively obtain important visual information and align them with textual features, so as to better instruct an LLM for CTRG.
Experiment results and further analysis on the benchmark M3D-Cap dataset show that our method outperforms strong baseline models and achieves state-of-the-art results, demonstrating its validity and effectiveness.
\end{abstract}

}

\maketitle
\IEEEdisplaynontitleabstractindextext
\IEEEpeerreviewmaketitle

\makeatletter
\def\@IEEEcompsocmakefnmark{\hbox{\normalfont\@thefnmark\ }}
\long\def\@makefntext#1{\parindent 1em\indent\hbox{\@IEEEcompsocmakefnmark}#1}
\makeatother

\makeatletter
\def\@IEEEcompsocmakefnmark{\hbox{\normalfont\@thefnmark.\ }}
\long\def\@makefntext#1{\parindent 1em\indent\hbox{\@IEEEcompsocmakefnmark}#1}
\makeatother

\renewcommand{\thefootnote}{\arabic{footnote}}

\section{Introduction}
\label{sec:introduction}

It is witnessed in the past several years that generating reports for chest X-ray radiographs attracts much attention \cite{chen-etal-2020-generating,hou2021ratchet,thawakar2024xraygpt},
where great success is observed in this area with satisfying performance.
Comparing to X-ray radiographs, generating reports for computed tomography (CT) images holds greater significance in clinical settings because radiologists need to process multiple images every time rather than a single or few radiographs for each case.
Although it is easy for one to borrow existing radiology report generation (RRG) models for CT reporting,
CT slices have their special characteristics, such as that images distribute along a spacial dimension, and changes across slices are highly continuous,
which brings unique challenges to the report generation process.
Existing studies for CT report generation (CTRG) follow the standard encoding-decoding paradigm that builds upon advanced visual encoders (e.g., convolutional neural networks or Transformer architectures) for extracting visual features from CT slices, and text generators (e.g., attention-based RNNs or Transformers) for producing the corresponding textual reports \cite{xie2019automated,zhou2019unet++,huang2020unet,mei2021sanet,mkindu2021lung,shi2021semi,liu2021medical,cao2022swin,aswiga2022multilevel,chen20243d,chen2024dia}.
For example, the M3D model \cite{bai2024m3d} extracts image features using 3D convolutional networks and integrates them with a text generator to produce diagnostic reports.
CT2Rep \cite{hamamci2024ct2rep} employs a cross-modal alignment to match visual features with texts.
To improve the performance of CTRG, there are studies that leverage extra sources, such as addition labeled data to perform multi-task processing \cite{kyung2024generative,di2024ct} and medical knowledge to align visual and textual features \cite{liu2021medical,van2022explainable,shi2023granularity,zhang2024co}.
Although some of these researches leverage large language models (LLMs) that are able to effectively generate textual content, the quality of the CT reports critically depends on accurate CT feature encoding and rigorous image-text alignments \cite{yang2023medxchat,thawakar2024xraygpt,park2024m4cxr}.
Existing methods are limited in fully accounting for the aforementioned characteristics of the images in CT volumes,
such as the incapability of
modeling the continuous correlations among CT slices as well as effectively integrating multi-level visual features.
Therefore,
it is expected to model visual features with a mechanism that appropriately treats such characteristics,
so that ensuring a more comprehensive representation of each CT volume.
By leveraging these crucial CT representations that carry essential information and are well-aligned with the text, LLMs should be more effective in understanding the semantics of the CT volume, and thus produce more accurate reports.

\begin{figure*}[t]
\centering
\includegraphics[width=0.95\textwidth, trim=0 10 0 0]{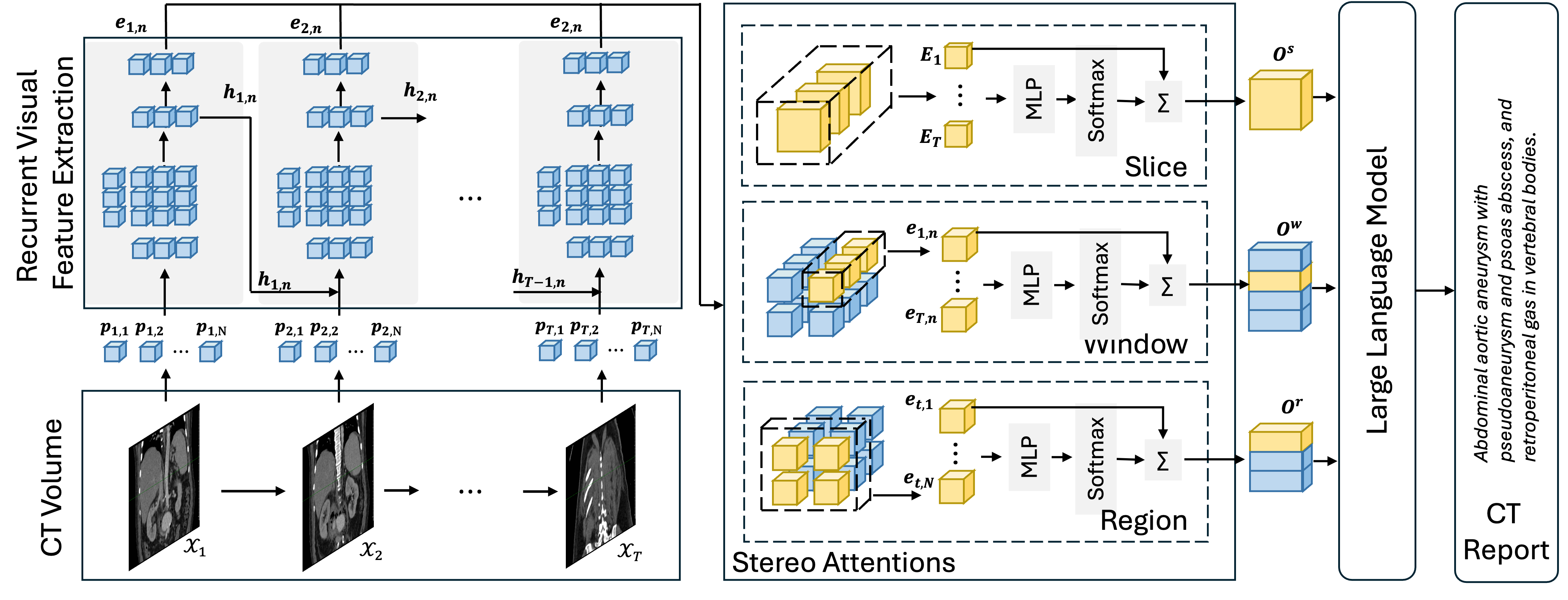}
\caption{
Illustration of our proposed method. The left part presents the input CT volume and the recurrent visual feature extraction process; the middle part shows the stereo attentions; and the right part displays report generation with an LLM.
}
\label{fig:framework}
\end{figure*}

In this paper, we propose a method for CTRG using LLM with recurrent visual feature extraction (R-VFE) for CT slice encoding and
stereo attentions to model essential information and align them across modalities.
Specifically, for R-VFE, we use a vision Transformer to recurrently encode CT slices from different positions, which is a novel encoding mechanism different from 3D CNN and 3D Transformers \cite{cciccek20163d,hatamizadeh2022unetr,cai2023swin}.
so that both historical and current CT slices are appropriately encoded with full consideration of their continuity.
Based on R-VFE, stereo attention weights the resulted visual features across multiple levels and aligns them with the textual feature space, 
so as to highlight the critical slices and regions containing pathologies that support accurate report content.
The final aligned representations are used to prompt the LLM to generate the CT report.
Experiments are performed on an English benchmark dataset M3D-Cap, and the results demonstrate the validity and effectiveness of our method, where it outperforms strong baselines and existing studies.

\section{Related Work}
\label{sec:related}

Generating textual reports for medical imaging, such as X-rays, CT scans, and MRIs, has long been a topic of significant interest for artificial intelligence application in medicine \cite{hou2021ratchet,lei2024autorg,shaker2024unetr++}.
Existing studies primarily focus on generating reports for chest X-ray images \cite{chen-etal-2020-generating,paul2021generalized,liu2023systematic,thawakar2024xraygpt,tian2024diffusion}, with relatively less attention given to CTRG and other types of medical imaging.
Existing studies on CTRG are primarily based on deep learning methods, converting imaging data into textual reports through visual feature extraction and text generation \cite{shi2021semi,aswiga2022multilevel,chen20243d},
where visual feature extraction typically relies on convolutional neural networks (CNNs) or Transformer-based models, while text generation is accomplished using language models based on LSTMs or Transformers.
For example, the M3D model \cite{bai2024m3d} utilizes 3D convolutional networks to extract features from CT slices and integrates them with a LLM (e.g., LLaMA-2 \cite{touvron2023llama}) to generate reports.
CT2Rep \cite{hamamci2024ct2rep} employs a cross-modal attention mechanism to align visual features with a Transformer-based text decoder.
Models such as UNETR++ \cite{shaker2024unetr++} and MTD-GAN \cite{kyung2024generative} enhance the representation of visual features through more efficient feature extraction architectures,
other studies such as Medical-VLBERT \cite{liu2021medical} introduce vision-language pre-trained models, combining visual and language features in a unified semantic space to improve the fluency and coherence of generated results.
To further improve CTRG, another line of existing research has explored incorporating additional optimization techniques and resources.
For example, CRHAN \cite{zhang2024co} leverages attention networks to encode co-occurrence relationships of lesions, extracting deeper semantic features.
Kyung et al. \cite{kyung2024generative} adopts multi-task learning framework to simultaneously optimize lesion detection and text generation.
Compared to existing studies, our method employs a hierarchical encoding strategy, utilizing a visual Transformer architecture to encode local information and attention mechanisms to identify and fuse important slices to capture the global spatial relationships in CT slices, enhancing CT slice representations and improving the effectiveness of report generation.

\section{Method}
\label{sec:method}

Our method performs an LLM-based CTRG with a novel encoding and hierarchical feature weighting mechanism.
The architecture of our method is presented in Figure \ref{fig:framework}, where there are three key components, namely, R-VFE, stereo attentions, and LLM for report generation.
In formally presenting our method,
given a CT volume, i.e., a sequence of consecutive CT slices, $\{\mathcal{X}_1, \mathcal{X}_2, \cdots, \mathcal{X}_T\}$,
where each $\mathcal{X}_t$ represents the slice at position $t$,
the R-VFE (denoted as $f_R$) encodes each position's slice and recursively incorporates the encoded representation into the input of the next position, resulting in a visual representation enriched with continuous information.
Subsequently, stereo attentions (denoted as $f_A$) are applied to R-VFE results from the slice, window, and region perspective to derive multi-level CT representations.
Finally, the resulted representations drive an LLM (denoted as $f_{LLM}$) to generate the report $\widehat{\mathcal{Y}}$,
with the overall method summarized by
\begin{equation}
    \widehat{\mathcal{Y}}
    = f_{LLM} \Bigl(\,
        f_{A}\bigl(
            f_R(\{\mathcal{X}_t\}_{t=1}^T)
        \bigr)
    \Bigr)
\end{equation}
The following texts provide details of our method along its pipeline.

\subsection{Recurrent Visual Feature Extraction}

In tackling the limitations of existing approaches that do not explicitly model the relationships among CT slices, especially the connections among the consecutive ones,
we propose R-VFE to encode the entire CT volume.
In detail,
our method employs a vision Transformer $f_{V}$ (e.g., a pre-trained ViT model \cite{dosovitskiy2020image}) as the base unit and recurrently uses it to encode the CT slices at different positions.
As illustrated in Figure \ref{fig:framework},
at each position $t$, the first step is to divide the CT slice $\mathcal{X}_t$ into a set of non-overlapping patches by
\begin{equation}
    x_{t,1}, x_{t,2}, \cdots, x_{t,N}
= f_{\text{patch}}\bigl(\mathcal{X}_t\bigr),
\end{equation}
where $N$ denotes the number of patches after division, and $x_{t,n}$ denotes the $n$-th patch in the $t$-th slice.
A linear mapping ($f_{lm}$) process is applied to obtain the corresponding patch embeddings $\mathbf{p}_{t,n}$ by $\mathbf{p}_{t,n} = f_{lm}(x_{t,n})$.
Then, the patch embeddings $\mathbf{p}_{t,n} $ are fused with the encoded patch features $\mathbf{h}_{t-1, n}$ from the previous position (i.e., $t-1$) and the standard position embedding $\mathbf{pos}_n$ by
\begin{equation} \label{eq:v}
    \mathbf{v}_{t,n} = \mathbf{p}_{t,n} + \mathbf{h}_{t-1, n} + \mathbf{pos}_n
\end{equation}
where the resulting $\mathbf{v}_{t,n}, (1\leq n \leq N)$ are fed into the vision Transformer $f_V$ to compute all patch features $\mathbf{h}_{t,n}, (1\leq n \leq N)$ for position $t$ by
\begin{equation} \label{eq:h}
    \mathbf{h}_{t,1} \cdots \mathbf{h}_{t,N} = f_{V} (\mathbf{v}_{t,1}, \cdots \mathbf{v}_{t,N})
\end{equation}
Particularly, if $t=1$, we use a zero vector as $\mathbf{h}_{t-1, n}$ for initialization.
Afterwards, a multi-layer perceptron (MLP) $f_{MLP}$ is applied to map $\mathbf{h}_{t,n}$ to the output patch representation $\mathbf{e}_{t,n} = f_{MLP} (\mathbf{h}_{t,n})$, where all patch representations are stacked to obtain the slice representation matrix $\mathbf{E}_{t} = [\mathbf{e}_{t,1}, \cdots, \mathbf{e}_{t,N}]$.
For the entire CT volume,
by iteratively performing Eq. (\ref{eq:v}) and Eq. (\ref{eq:h}), the information of both historical and current slices are encoded and passed to the encoding process of the next slice 
until we reach the last position $T$,
where all slice and patch representations throughout the process are used for later processing.

\subsection{Stereo Attentions}

Although R-VFE encodes continuous features from CT volumes and captures slice-wise continuity, it is still required to identify crucial information which is specific to local lesions, key tissue structures, or pathological changes,
thus emphasizing on the features that significantly contribute to the final report.
Considering attention mechanism is demonstrated to be an effective approach to identify important content \cite{vaswani2017attention,Huang2019AttentionOA,tian-etal-2020-supertagging,zhang2024co,tian2024learning}.
Therefore, we propose stereo attentions, including three attentions from different perspectives that identify important patches at slice, window, and region levels,
and selectively fuses them to align with the textual features, so as to facilitate cross-modal content processing.
The details are illustrated as follows.

\subsubsection{{Slice Attention}}

This attention applies to
the entire volume to select the slices
most likely to contain critical information.
In doing so, we use a matrix flatten operation and a fully connected layer $f_{s}$ to map the slice representation $\mathbf{E}_{t}$ to a scalar score and then employ a softmax to obtain the attention weights $\alpha_{t}$.
Then, we compute a weighted sum over all matrix $\mathbf{E}_{t}$ to obtain a slice-level feature matrix, which is passed through a fully connected layer $g_s$ to obtain $\mathbf{O}^s$ that aligns with the textual semantic space of the LLM,
formulated by
\begin{equation}
    \mathbf{O}^s
    = g_s (\sum_{t=1}^{T} \alpha_{t} \cdot \mathbf{E}_{t}), 
\end{equation}
where $\alpha_{t}$ is computed by
\begin{equation}
    \alpha_{t}
    = \frac{\exp\bigl(f_{s}( f_f ({\mathbf{E}}_{t}))\bigr)}{\sum_{t=1}^{T} \exp\bigl(f_{s}( f_f ({\mathbf{E}}_{t}))\bigr)}
\end{equation}

\subsubsection{{Window Attention}}

This attention weight the patches at the fixed positions across slices,
which determines the relative importance of the features at the same position over different slices, aiding in the identification of lesions.
Specifically, for the $n$-th patch, we map its representation to a scalar score using a fully connected layer \(f_w\),
with a softmax function applied to compute the cross-slice weight \(\beta_{t,n}\).
Next, we apply \(\beta_{t,n}\) to \(\mathbf{e}_{t,n}\) to compute the cross-slice weighted sum, obtaining the representation for the \(n\)-th patch.
Similar to slice attention, the output is computed through a similar process by
\begin{equation}
    \mathbf{o}^w_p 
    = g_w (\sum_{t=1}^{T} \beta_{t,n} \cdot \mathbf{e}_{t,n} )
\end{equation}
where $\beta_{t,n}$ is computed by
\begin{equation}
        \beta_{t,n} 
    = \frac{\exp\bigl(f_{w}(\mathbf{e}_{t,n})\bigr)}
           {\sum_{t=1}^{T}\exp\bigl(f_{w}(\mathbf{e}_{t,n})\bigr)}
\end{equation}
We perform this operation for all patches, resulting in \(\mathbf{o}^w_1, \cdots, \mathbf{o}^w_N\), which are stacked to obtain the matrix $\mathbf{O}^w=[\mathbf{o}^w_1, \cdots, \mathbf{o}^w_N]$.

\begin{table}[t]
\centering
\caption{
The statistics of the datasets used in the experiments, where the number of CT slices, reports, as well as the average number of tokens are reported.
}
\begin{tabular}{l|ccc}
\toprule
 & \textbf{Train} & \textbf{Dev} & \textbf{Test} \\
\midrule
\# of Slices 
& 8,042,725 & 224,891 & 171,556 \\
\# of Reports 
& 116,092 & 2,000 & 2,000 \\
\# of Tokens 
& 7,034,044 & 142,108 & 134,084 \\
\midrule
Avg. \# of Slices 
& 69.3 & 122.4 & 85.8 
\\
Avg. \# of Tokens
& 60.6 & 70.1 & 67.0
\\
\bottomrule
\end{tabular}
\label{tab:dataset}
\end{table}

\subsubsection{{Region Attention}}

This attention focuses on the important regions within each slice.
In doing so,
for the $t$-th slice with feature matrix $\mathbf{E}_t = [\mathbf{e}_{t,1}, \cdots, \mathbf{e}_{t,N}]$, we employ a fully connected mapping $f_{r}$ to produce scores for each patch and apply a softmax function to obtain the attention weights $\gamma_{t,n}$.
Subsequently, we compute a weighted sum of the patch vectors, obtaining the final region-level representations.
Similar to the slice and window attention, we utilize a fully connected layer $g_r$ to map $\mathbf{a}^r_t$ to \(\mathbf{o}^{r}_{t}\) through
\begin{equation}
    \mathbf{o}^r_t 
    = g_r (\sum_{n=1}^{N} \gamma_{t,n} \cdot \mathbf{e}_{t,n})
\end{equation}
where $\gamma_{t,n}$ is computed by
\begin{equation}
    \gamma_{t,n} 
    = \frac{\exp\bigl(f_{r}(\mathbf{e}_{t,n})\bigr)}
           {\sum_{p=1}^{N}\exp\bigl(f_{r}(\mathbf{e}_{t,n})\bigr)}
\end{equation}
Similarly, upon obtaining $\mathbf{o}^r_1 \cdots \mathbf{o}^r_T$ within a slice, we stack them to obtain the matrix $\mathbf{O}^r=[\mathbf{o}^r_1, \cdots, \mathbf{o}^r_T]$.

\begin{table*}[t]
\centering
\caption{Results of baselines and our method on the test set of M3D-Cap. 
SA refers to stereo attentions; BL, RG, MT, and B-S stand for BLEU, ROUGE-1, METEOR, BERT-score, respectively.}
\vspace{-0.2cm}
\begin{tabular}{l | ccccccc}
\toprule
 & BL-1 & BL-2 & BL-3 & BL-4 & RG & MT & B-S \\
\midrule
Qwen2-VL 
& 13.27 & 5.47 & 3.05 & 1.03 & 15.46 & 11.83 & 87.41 \\
\ + R-VFE
& 16.43 & 7.73 & 3.21 & 1.27 & 18.74 & 14.46 & 88.21 \\
\ + R-VFE + SA (Ours)
& 17.03 & 8.12 & 3.40 & 1.42 & 19.21 & 14.63 & 88.67  \\
\bottomrule
\end{tabular}
\vspace{-0.1cm}
\label{tab:main_results}
\end{table*}

\begin{table}[t]
\centering
\caption{Comparison of our method with existing representative studies.}
\vspace{-0.2cm}
\begin{tabular}{l | cccc}
\toprule
 & BLEU & ROUGE & METEOR & BERT-score \\
\midrule
RadFM \cite{wu2023towards}
& 12.23 & 16.49 & 11.57  & 87.93 \\
M3D \cite{bai2024m3d}
& 15.15 & \textbf{19.55} & 14.38 & 88.46 \\
\midrule
Ours 
& \textbf{17.03} & 19.21 & \textbf{14.63} & \textbf{88.67}  \\
\bottomrule
\end{tabular}
\label{tab:sota}
\end{table}

\subsection{Report Generation}

Upon the collection of the results, $\mathbf{O}^s$, $\mathbf{O}^w$, and $\mathbf{O}^r$, from stereo attentions over the slice-, window-, and region-level features,
we combine them to feed into an LLM for report generation.
Specifically, we firstly use a fully connected layer $g$ to transform $\mathbf{O}^s$, $\mathbf{O}^w$, and $\mathbf{O}^r$ into prompt embeddings compatible with the LLM by
\begin{equation}
    \mathbf{Z} 
    = 
    \bigl[
        g(\mathbf{O}^s),
        g(\mathbf{O}^w),
        g(\mathbf{O}^r)
    \bigr]
\end{equation}
So that features at different levels are preserved and serve as ``soft prompts'' to instruct the decoding process of the LLM.
Then, we feed the prompt embeddings $\mathbf{Z}$ into $f_{LLM}$ and autoregressively generate the final report $\widehat{\mathcal{Y}}$.
During training, we minimize the cross-entropy loss between the generated report $\widehat{\mathcal{Y}}$ and the gold standard report ${\mathcal{Y}}^*$, thereby updating all parameters in the visual encoder, the attention modules, and the LLM layers.

\section{Experiment Settings}
\label{sec:exp_settings}

\subsection{Dataset}

In our experiments, we utilize the M3D-Cap \cite{bai2024m3d} dataset,
which focuses on CT scans of various body parts, including the chest, abdomen, and pelvis.
This dataset is designed to support tasks that require cross-modal understanding, combining high-resolution CT slices with corresponding reports. 
We use the standard train/dev/test split \cite{bai2024m3d}, where there are 116,092, 2,000, and 2,000 instances (one instance contains an entire CT volume with its report) in the training, development, and test datasets, respectively.
The statistics of the dataset is reported in Table \ref{tab:dataset}.

\subsection{Implementation Details}

Existing studies show that a high-quality input representation plays an essential role in satisfying model performance \cite{mikolov2013efficient,song-etal-2017-learning,song2018joint,devlin-etal-2019-bert,lewis-etal-2020-bart,touvron2023llama,li2024llava,tian-etal-2024-chimed}.
Therefore, in our experiments, we utilize Qwen2-VL (2B) \cite{wang2024qwen2}\footnote{We obtain the model from \url{https://huggingface.co/Qwen/Qwen2-VL-2B-Instruct}.}, which achieves remarkable performance in many downstream tasks \cite{liu2025visual,tian2025representation,zhang2025r1}, as the visual encoder and the LLM,
with its original visual encoder as the basic unit for R-VFE to process each CT slice and utilize the LLM as the report generator.
We follow the default settings of Qwen2-VL, where the visual encoder has 32 layers of Transformers with a hidden size of 1280 and the LLM contains 28 Transformer layers with a hidden size of 1,536.
To ensure proper processing of the images by the visual encoder, we resize each image to the dimension of 224 × 224.
During training, we update all parameters of both the visual encoder and the LLM.
We train the model for 3 epochs with a learning rate of 5e-5,
with the batch size set to 16, and weight decay applied with a coefficient of 0.01.
For evaluation, we adopt standard natural language generation (NLG) metrics, including BLEU, ROUGE, METEOR, and BERT-score.

\begin{figure}[t]
    \centering
    \includegraphics[width=0.5\textwidth, trim=0 20 0 0]{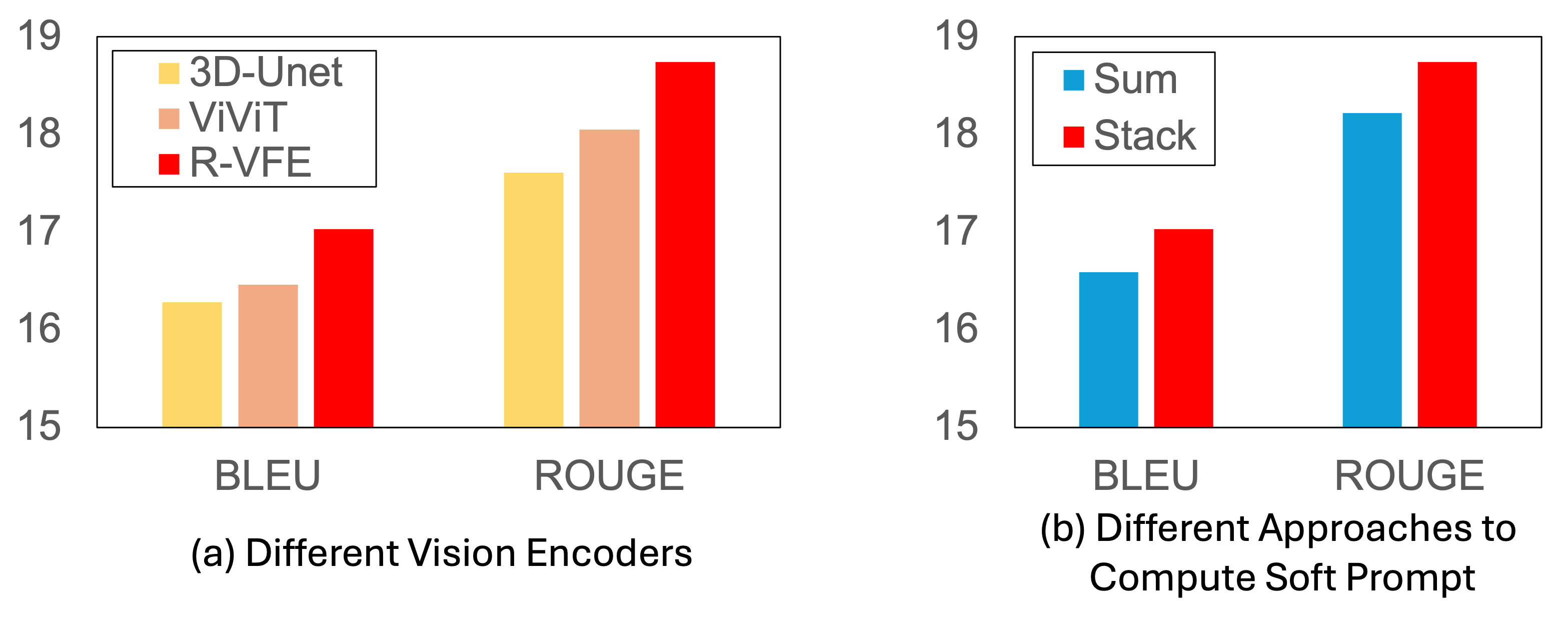} 
    \caption{
    The BLEU-1 and ROUGE-1 scores of different methods. (a) shows the results of different modules to encode the CT volume; (b) presents the results of different strategies for computing the soft prompt.
    }
    \label{fig:recurrent}
\end{figure}

\begin{table*}[t]
\centering
\setlength{\tabcolsep}{6pt} 
\caption{Results of ablation study on different attentions, where the slice-, window-, and region-level attention is ablated.}
\vspace{-0.2cm}
\begin{tabular}{c c c | ccccccc }
\toprule
Slice & Window & Region & BL-1 & BL-2 & BL-3 & BL-4 & RG & MT & BS\\
\midrule
$\checkmark$ & $\checkmark$ & $\checkmark$
& 17.03 & 8.12 & 3.40 & 1.42 & 19.21 & 14.63 & 88.67  \\
\midrule
$\checkmark$ & $\times$ & $\times$
& 16.60 & 7.82 & 3.27 & 1.31 & 18.95 & 14.49 & 88.35\\
$\times$ & $\checkmark$ & $\times$
& 16.78 & 7.94 & 3.32 & 1.34 & 19.03 &14.57 & 88.49 \\
$\times$ & $\times$ & $\checkmark$ 
& 16.65 & 7.78 & 3.23 & 1.30 & 18.87 & 14.48 & 88.29 \\
\midrule
$\times$ & $\times$ & $\times$
& 16.43 & 7.73 & 3.21 & 1.27 & 18.74 & 14.46 & 88.21 \\
\bottomrule
\end{tabular}
\label{tab:ablation}
\vspace{-0.3cm}
\end{table*}

\section{Results and Analysis}
\label{sec:results}

\subsection{Overall Results}

We compare our method with two baselines.
The first one uses a standard multimodal LLM (denoted as ``Qwen2-VL''), where each CT slice is encoded individually, and the resulting representations of all slices are concatenated and fed into the LLM to generate the corresponding report;
the second one (denoted as ``Qwen2-VL + R-VFE'') adds R-VFE on top of Qwen2-VL.
The results of the baselines and our method are presented in Table \ref{tab:main_results},
with several observations.

First, compared with the Qwen2-VL baseline, the model with R-VFE achieves better performance across multiple evaluation metrics,
which indicates the benefit of explicitly modeling the connections among continuous CT slices.
Second, our approach with both R-VFE and attentions achieves further improvements over the ``Qwen2-VL + R-VFE',
which illustrates that our attentions are able to effectively identify important information and align them with the textual ones, so as to help the LLM to produce better reports.
In addition, we compare our method with existing studies, and report the results in Table \ref{tab:sota}.
It is observed that our results achieve the state-of-the-art performance over all studies, including M3D \cite{bai2024m3d} that utilizes LLaMA-2 7B as the LLM with more parameters to generate the report.
This observation highlights the importance of the proposed R-VFE and stereo attentions for CTRG.

\begin{figure*}[t]
    \centering
    \includegraphics[width=0.9\textwidth, trim=0 0 0 0]{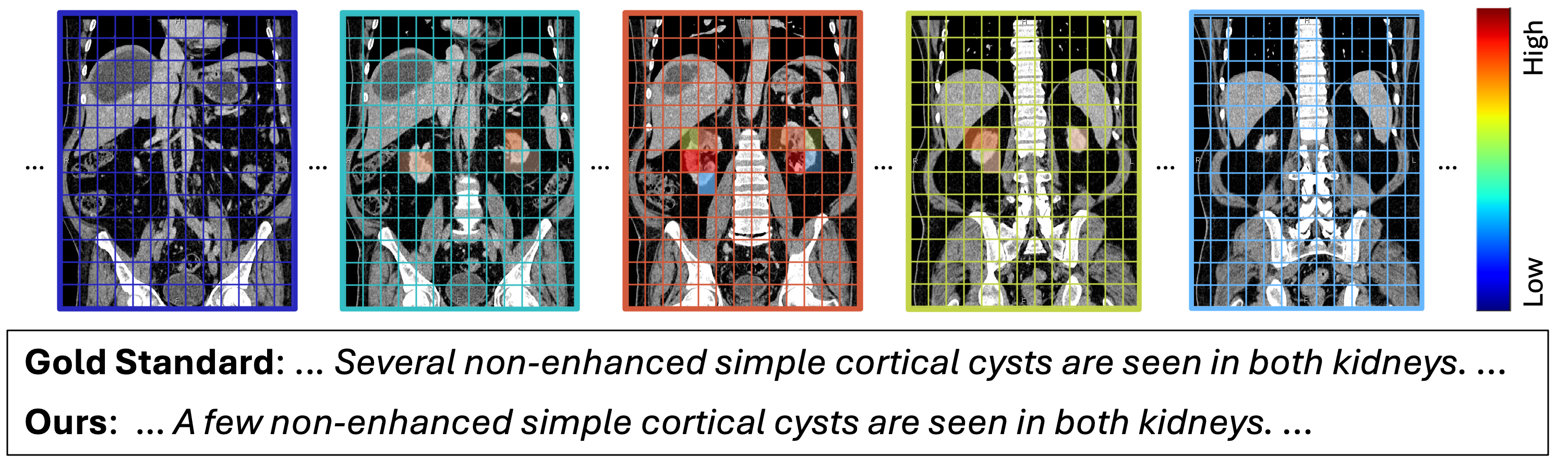} 
    \caption{
    A case study with several input CT slices, a part of the gold standard, and the output of our method.
    The grid denotes the boundary of different patches in a slice; the slice attention is visualized through the color of the border and grid of each slice; the window attention is visualized through the colors in the grid patches.
    }
    \label{fig:case_study}
    \vspace{-0.2cm}
\end{figure*}

\subsection{Analysis}

To analyze the validity of recurrent modeling for visual features, we run our approach with other types of 3D vision encoders, namely, 3D-Unet 
\cite{cciccek20163d} and ViViT \cite{arnab2021vivit}, whose output is fed into the attention mechanism and the LLM as that R-VFE does.
We report the results (i.e., the BLEU-1 and ROUGE-1) of these models on M3D-Cap in Figure \ref{fig:recurrent}(a), where
it is observed that, our method with the recurrent design achieves the best performance, indicating the effectiveness of our approach since it is able to model the information across continuous slices.

Similarly, we investigate different strategies to leverage the output of three attentions.
Specifically, we run experiments where the prompt embedding $\mathbf{Z}$ is the sum of all attention output (i.e., $\mathbf{Z}=g(\mathbf{O}^s)+g(\mathbf{O}^w)+g(\mathbf{O}^r)$)
The results of this strategy, as well as our method, are reported in Figure \ref{fig:recurrent}(b).
We observe that our method achieves better performance than the ``Sum'' strategy, which is explained by that, the ``Sum'' strategy fuses the features and thus has the potential to miss information of particular individual attention output, whereas our strategy stacks the attention output with less possible information loss.

To analyze the contributions of slice, window, and region features, we test if only using one of them.
The results are shown in Table \ref{tab:ablation}, which also includes the performance of our full model and the ``Qwen2-VL + R-VFE'' baseline for comparison.
As observed, using one of the three attentions improves the performance of the baseline,
where the window attention presents the highest improvements,
because pathological areas in a CT volume generally present distinct characteristics.
This difference is particularly evident across consecutive CT slices.  
For example, in regions that are expected to exhibit higher gray levels, low-gray-level features appear and gradually increase-in-size across slices, suggesting that disease is possibly presented.
In such cases, the window attention that integrates cross-slice attention targeting specific regions is able to detect these abnormal features, thereby assisting in the identification of pathological areas.
In contrast, attention mechanisms focusing only on slices or regions may spread across multiple organs, leading to insufficient process on particular evident areas.

To perform qualitative analysis,
Figure \ref{fig:case_study} presents a case study with several CT input slices, the gold standard, the output of our method, and the visualization of slice and window attentions  (the region attentions are ommitted because they are similar to the window attention). 
We observe that our method correctly identifies the cortical cysts in both kidneys in the report, assigns high weights to the relevant slice (i.e., the third one) and the regions that are relevant to the disease, which accordingly enhances the interpretability of the generated report.

\section{Conclusion}
\label{sec:conclusion}

In this paper, we propose an LLM-based method for CTRG with recurrent encoding and stereo attention.
Specifically, we encode CT slices in a recurrent manner and then perform a comprehensive attention mechanism over all visual features to distinguish the important information for CTRG at different levels and align visual features with textual ones.
Finally we use LLM to accommodate the extracted and aligned features and generate the report.
Results on an English benchmark dataset for CTRG (i.e., M3D-Cap) validate the effectiveness of our method, which outperforms strong baselines and achieves state-of-the-art results.
Analyses also confirm that our method is able to highlight important slices and regions that are relevant to important pathologies and thus help CTRG.

\nocite{langley00}

\ifCLASSOPTIONcaptionsoff
  \newpage
\fi

\bibliographystyle{IEEEtran}
\bibliography{reference}

\newpage

\appendices

\end{document}